%% file: root.tex
\documentclass[letterpaper, 10 pt, conference]{ieeeconf}  

\IEEEoverridecommandlockouts                              

\usepackage{amsmath} 
\usepackage{amssymb}  

\usepackage{graphicx} 
\usepackage{array}

\newlength{\bibitemsep}
\newlength{\bibparskip}
\let\oldthebibliography\thebibliography
\renewcommand\thebibliography[1]{%
	\oldthebibliography{#1}%
	\setlength{\parskip}{\bibitemsep}%
	\setlength{\itemsep}{\bibparskip}%
}

\graphicspath{{./pdf/}{./jpg/}}
\DeclareGraphicsExtensions{.pdf,.jpg,.png}

\title{\LARGE \bf
FDMO: Feature Assisted Direct Monocular Odometry
}

\author{Georges Younes$^{1,2}$, Daniel Asmar$^{2}$ and John Zelek$^{1}$
\thanks{$^{1}$System Design Department in the Faculty of Engineering, University of Waterloo, 200 University Ave W, Waterloo, Canada.}
\thanks{$^{2}$Mechanical Engineering Department in the Faculty of Engineering and Architecture, American University of Beirut, Riad El Solh, Beirut, Lebanon.}}

\begin{document}

\maketitle
\thispagestyle{empty}
\pagestyle{empty}

\begin{abstract}
	Visual Odometry (VO) can be categorized as being either direct or feature based. When the system is calibrated photometrically, and images are captured at high rates, direct methods have shown to outperform feature-based ones in terms of accuracy and processing time; they are also more robust to failure in feature-deprived environments.  On the downside, Direct methods rely on heuristic motion models to seed the estimation of camera motion between frames; in the event that these models are violated (e.g., erratic motion), Direct methods easily fail.  This paper proposes a novel system entitled FDMO (Feature assisted Direct Monocular Odometry), which complements the advantages of both direct and featured based techniques.  FDMO bootstraps indirect feature tracking upon the sub-pixel accurate localized direct keyframes only when failure modes (e.g., large baselines) of direct tracking occur.  Control returns back to direct odometry when these conditions are no longer violated. Efficiencies are introduced to help FDMO perform in real time.  FDMO shows significant drift (alignment, rotation \& scale) reduction when compared to DSO \& ORB SLAM when evaluated using the TumMono and EuroC datasets. 
	

\end{abstract}

\input{introduction}

\input{related_work}

\input{proposed_system}

\input{experiments}

\input{conclusions}







\section*{ACKNOWLEDGMENT}
This work was funded by 
the University Research Board (UBR) at the American University of Beirut, and the Canadian National Science Research Council
(NSERC).


\bibliographystyle{IEEEtran}
\bibliography{younes_iros}


\end{document}

%% file: introduction.tex
\section{INTRODUCTION}
\label{sec:intro}
Visual Odometry (VO) is the process of localizing one or several cameras in an unknown environment. Using a video feed from a moving camera,
VO  generates a temporary 3D map of the camera's surroundings and uses it to recover the camera's motion within the observed scene.
 VO is considered indispensable
for various tasks, including visual-based robotic navigation and augmented reality applications, to name a few.
	
Two decades of extensive research have led to a multitude of VO systems that can be categorized
based on the type of information they extract from an image, as direct, feature-based, or a hybrid of
both \cite{younes_2016_ARXiv}. 
While the direct framework manipulates photometric measurements (pixel intensities), the feature-based framework extracts and uses visual features as an intermediate image representation. 
The choice of feature-based or direct method has important ramifications on the performance of the entire VO system, with each type exhibiting its own challenges, advantages, and disadvantages. 
\begin{figure}[!tb] 
		\centering
		\includegraphics[trim={0.2cm 0.3cm 0 0.7cm},clip,width=0.48\textwidth]{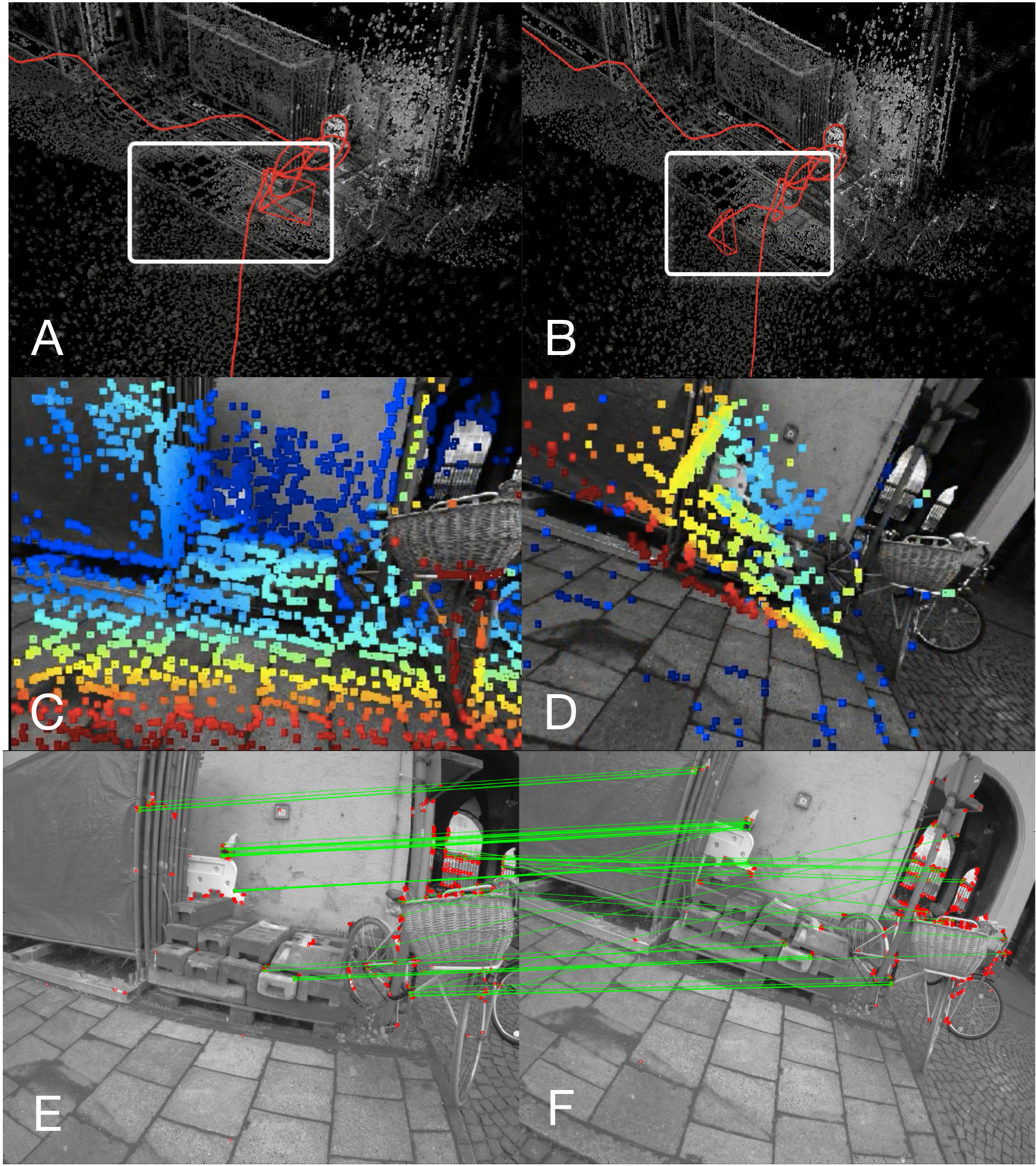}
		\vspace*{-0.7cm}
        \caption{Direct methods failure under large baseline motion.
        (A) and (B) show the trajectory estimated from a direct odometry system, before and after going through a relatively large baseline between two consecutive frames (shown in (C) and (D). Notice how the camera's pose in (B) derailed from the actual path to a wrong pose. (C) and
        (D) show the projected direct point cloud on both frames respectively after erroneously
        estimating their poses. Notice how the projected point cloud is no longer aligned with the scene. On the other hand, (E) and (F) show how features can be matched across the relatively large baseline, allowing feature-based methods to cope with such motions.}
		\label{fig:motivation}
	\vspace{-0.62cm}
	\end{figure}
One disadvantage of particular interest to this paper is the sensitivity of direct methods to their motion model. This limitation is depicted in Fig. \ref{fig:motivation} (A) and (B), where a direct VO system is subjected to a motion that violates its presumed motion model, and causes it to erroneously expand the map as shown in Fig. \ref{fig:motivation} (C) and (D). 
Inspired by the invariance of feature-based methods across relatively large baselines (as shown in Fig. \ref{fig:motivation} (E) and (F)), 
this paper proposes to address the shortcomings of direct methods, by detecting failure in their frame to frame odometry component, and accordingly invoking an efficient feature-based strategy to cope with the large baselines. We call our approach Feature assisted Direct Monocular Odometry, or FDMO for short.
While we don't make use of a complete
SLAM formulation (\textit{i.e.}, we don't make use of a global map for failure recovery nor for loop closure), we show that by effectively exploiting information available from the direct framework, FDMO inherits the advantages of direct methods in terms of sub-pixel accuracy, robustness to feature deprived environments in its feature-based map, and low computational cost at frame rate; all while gaining the advantages of the feature-based framework in terms of handling large baseline motions.


\section{Background}
\label{sec:Backg}
Visual odometry can be broadly categorized as being either direct or feature-based.
\subsection{Direct VO}
\label{sec:dir}
Direct methods process raw pixel intensities with
the brightness constancy assumption \cite{baker_2004_IJCV}:
\begin{equation}
\label{eq:brightness}
I_{t}(x)=I_{t-1}(x+g(x)),
\end{equation}
where $x$ is the 2-dimensional pixel
coordinates $(u,v)^T$ and $g(x)$ denotes the displacement function of $x$ between the two images $I_t$and $I_{t-1}$.
\subsubsection{Traits of direct methods}
since direct methods rely on the entire image for localization, they are less susceptible to failure
in feature-deprived environments, and do not require a time-consuming feature extraction and matching step.  More importantly, since the alignment takes place at the pixel
intensity level, the photometric residuals can be interpolated over the image domain $\Omega I$,
resulting in an image alignment with sub-pixel accuracy, and relatively less drift than feature-based odometry methods \cite{irani_1999_iccv}.  
However, the objective function to minimize is highly non-convex; its convergence basin is very small, and will lock to
an erroneous configuration if the optimization is not accurately initialized.  Most direct methods
cope with this limitation by adopting a pyramidal implementation, by assuming small inter-frame
motions, and by relying on relatively high frame rate cameras; however, as a rule of thumb, all
parameters involved in the optimization should be initialized such that $x$ and
$g(x)$ are within 1-2 pixel radii from each other.  
\subsubsection{State of the art in direct methods}
Direct Sparse Odometry (DSO) \cite{engel_2016_ARXIV} 
is a keyframe-based VO that adopts the inverse depth parametrization
of \cite{civera_2008_TRO}, which is suitable for estimating depths with small parallax; 
therefore, it does not suffer from epipolar-geometry-based triangulation degeneracies 
and can handle points at infinity.
DSO employs a pyramidal implementation of the forward additive image alignment \cite{baker_2004_IJCV}
to optimize a variant of the brightness constancy assumption over the incremental
geometric transformation between the current frame and a reference keyframe. The optimization can be
summarized by:
\begin{equation}
\label{eq:FAIA}
\underset{T_{f_i,KF_j}}{\operatorname{argmin}}\sum_{x} \sum_{x_k\in N(x)}Obj(I_{f_i}(\omega(x_k,d,T_{f_i,KF_j})-I_{KF}(x_k,d)))
\end{equation}
where $T_{f_i,KF_j} \in SE(3)$ is the transformation relating the current frame $f_i$ to a
reference keyframe $KF_j$; $x\in \Omega I_d$ is the set of image locations with sufficient intensity
gradient and an associated depth value $d$. $x_k\in N(x)$ is the set of pixels surrounding $x$
defined by the local neighborhood $N(x)$. Obj(.) is the Huber norm, and $\omega(.)$ is defined as:
\begin{equation}
\omega(x_k,d,T_{f_i,KF_j})=\pi(T_{f_i,KF_j}\pi^{-1}(x_k,d))
\end{equation}
DSO's tracking front-end takes place on a frame by frame basis, and exploits the nature of small
inter-frame motions in a video feed to update its depth filters for each point of interest
in a keyframe, as described in \cite{engel_2013_ICCV}.
DSO keeps in its map a small set of keyframes $\kappa_{dir}$, in which all current map points exist.
DSO's back-end ensures the local consistency of its map through a photometric optimization, defined by:
\begin{multline}
\label{eq:PhBA}
\underset{T_{KF_i},d}{\operatorname{argmin}}\sum_{KF_i\in \kappa_{dir}}\sum_{x}\sum_{\substack{j\in \kappa_{dir}\\ i\neq j}} \sum_{x_k\in N(x)}\\Obj(I_{KF_i}(\omega(x_k,d,T_{KF_i,KF_j})-I_{KF_j}(x_k,d)))
\end{multline}

\subsection{Feature-based VO}\label{sec:feat}
Feature-based methods process 2D images to extract locations that are salient in an image. Let $x=(u,v)^T$ represent a feature's pixel coordinates in the 2-dimensional image domain
$\Omega\textit{I}$. Associated with each feature is an $n$-dimensional vector $Q^{n}(x)$, known as a
\textit{descriptor}.  The set $\Phi\textit{I}\{x,Q(x)\}$ is an intermediate image representation
after which the image itself becomes obsolete and is discarded.
\subsubsection{Traits of feature-based methods}
on the positive side, features with their associated descriptors are somewhat invariant to viewpoint
and illumination changes, such that a feature $ x\in\Phi I_1$ in one image can be identified as
$x'\in \Phi I_2$ in another, across \textit{relatively large baselines}. Such invariance comes from the properties of a feature extractor. 
On the downside, and as a result of their discretized image representation space, feature-based solutions offer inferior accuracy when compared to 
\textit{direct} methods, where the image domain can be interpolated for sub-pixel accuracy.
\subsubsection{State of the art in feature-based methods}
ORB-SLAM \cite{mur-artal_2015_TRO}, currently considered the state of the art in feature-based
methods, associates FAST corners \cite{rosten_2006_ECCV} with ORB descriptors \cite{rublee_2011_ICCV}
as an intermediate image representation. The ORB SLAM map consists of 3D map points
$X_j(\{x_i,Q(x_i)\}) \in {\rm I\!R}^3$, as well as special frames, referred to as \textit{keyframes}
(KF), where $KF_i\in\kappa=\left[
T_{i,w}, \Phi\{x,Q(x)\} \right]$ with $T_{i,w}\in SE(3)$ being the keyframe's pose in the world
coordinate frame.
The 3D points are triangulated using Epipolar geometry
\cite{hartley_2003_Cambridge}, from multiple observations of the feature $\{x_i,Q(x_i)\})$ in two or
more keyframes.  Unfortunately,
this adds another shortcoming to feature-based methods, as Epipolar based
triangulation is unstable for "far-away" features (small
parallax) \cite{yang_2017_Arxiv}. 

Regular frames are localized by minimizing the geometric re-projection error defined by:
\begin{equation}
	\label{eq:featTracking}
	\underset{T_{i}}{\operatorname{argmin}} \sum_j Obj(x_j-\pi(T_{i,w},X_j)),
\end{equation}
where Obj(.) is the Huber norm, $x_j$ is the 2D location, in the current frame, of the feature that
matched the 3D point $X_j$; $\pi$ is the pinhole camera projection model. 
3-dimensional point $X_j$ onto the current frame. 
The consistency of the map is maintained through a local bundle adjustment process defined by:
\begin{equation}
	\underset{T_{KF_i},X_j}{\operatorname{argmin}} \sum_{i\in\kappa'}\sum_{j}
    Obj(x_{i,j}-\pi(T_{KF_i},X_j)),
\end{equation}
where $X_j$ is the set of map points that were observed in the set of keyframes $KF_i \in \kappa'$
and $\kappa'$ is a subset of the map. 
 Both optimizations are resilient to relatively large inter-frame baseline motions and have a large convergence radius.
	

Although ORB SLAM is considered a SLAM system (which maintains a global map and uses it for loop closure), its VO component is considered the state of the art in feature-based methods. Therefore, for the fairness of comparison, and similar to \cite{mono_dataset}, we reduce ORB SLAM to an odometry system by disabling its loop closure detection component.

\subsection{Feature-based vs. Direct}
\label{sec:hybrid}
When the corresponding pros and cons of both feature-based and direct frameworks are placed side by
side, a pattern of complementary traits emerges (Table \ref{tab:FeatvsOdom}). An ideal framework would exploit both direct and feature-based advantages to benefit from
the direct formulation accuracy and robustness to feature-deprived scenes, while making use of feature-based methods for large
baseline motions.
\begin{table}[!htb]
	\caption{Comparison between the feature-based and direct methods. The more of the symbol +, the
    higher the attribute.}
	\label{tab:FeatvsOdom}
	\begin{center}
        \begin{tabular}{|p{3.5cm}||c||c|}
	\hline
    \textbf{Trait} & \textbf{Feature-based} & \textbf{Direct}\\
	\hline
	Large baseline & +++ & +\\
	\hline
	Robust to Feature Deprivation& + & +++\\
	\hline
	Recovered scene point density  &+ & +++\\
	\hline
	Accuracy & + & +++\\
	\hline
	Optimization Non-Convexity& + & ++\\
	\hline
\end{tabular}
\end{center}
\vspace*{-0.5cm}
\end{table}

%% file: related_work.tex
\section{Related work}
\label{sec:related}
Hybrid direct-feature-based systems were previously proposed in \cite{forster_2014_ICRA},
\cite{krombach_2016_ICIAS} and \cite{jellal_2016_ECMR}; however, \cite{forster_2014_ICRA} did not
extract feature descriptors, it relied on the direct image alignment to perform data association
between the features. While this led to significant speed-ups in the processing required for data association, it could not handle large baseline motions; as a result, their work was limited to high frame rate cameras (which ensured frame-to-frame motion is small).
On the other hand, both \cite{krombach_2016_ICIAS} and \cite{jellal_2016_ECMR} adopted a
feature-based approach as a front-end to their system, and subsequently optimized the measurements
with a direct image alignment.  As such, these systems suffer from the limitations of the
feature-based framework and are subject to failure in feature-deprived environments. To address this issue, both systems resorted to stereo cameras. In contrast to these systems, we propose a direct alignment as a front-end,
backed by a feature-based map that is invoked whenever the direct alignment fails. Therefore, FDMO can operate using a monocular camera, and can adaptively switch between the two modes when necessary. It is it noteworthy to mention that FDMO can be adapted for stereo and RGBD cameras as well.

%% file: proposed_system.tex
\begin{figure*}[!htb] 
	\centering
	\includegraphics[width=0.7\textwidth]{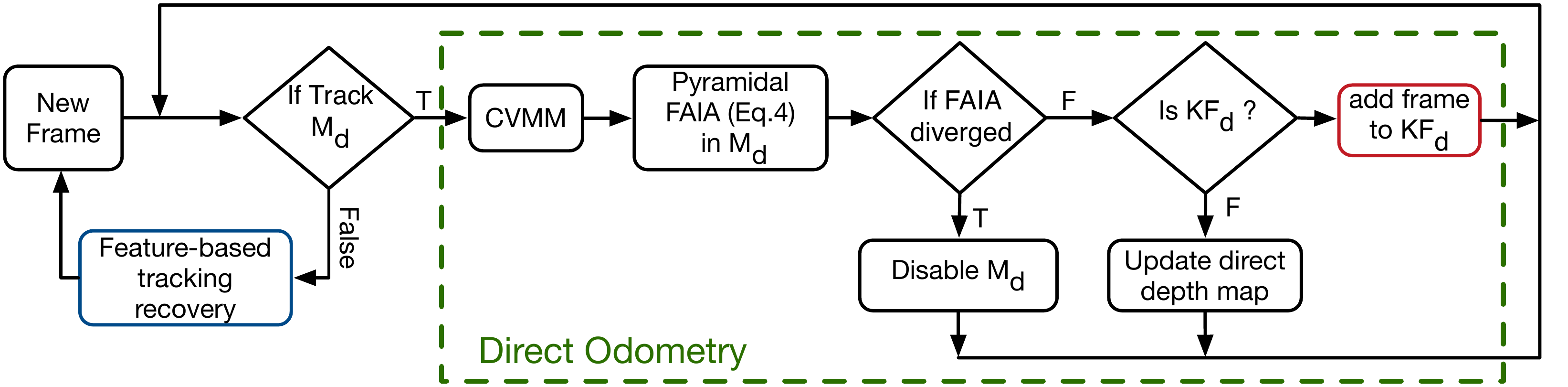}
	\caption{Front-end flowchart of FDMO. It runs on a frame by frame basis and uses a constant velocity motion model (CVMM) to seed a Forward Additive Image Alignment (FAIA) to estimate the new frame's pose and update the direct map depth values. It also decides whether to invoke the feature-based tracking or add a new keyframe into
		the system.}
	\label{fig:DirAlignment}
\end{figure*}
\section{Proposed system}
	\label{sec:proposed}
To capitalize on the advantages of both feature-based and direct frameworks, our proposed approach consists of a local direct visual odometry, assisted with a feature-based map, such that it may resort to feature-based odometry only when necessary. Therefore, FDMO does not need to perform a computationally expensive feature extraction and matching step at every frame. During its feature-based map expansion, FDMO exploits the localized keyframes with sub-pixel accuracy from the direct framework, to efficiently establish feature matches in feature-deprived environments using restricted epipolar search lines.

To address any ambiguities, the subscript
\textit{d} will be assigned to all direct-based measurements and \textit{f} for feature-based
measurements.  
Similar to DSO, FDMO's local
temporary map $M_{d}$ is defined by a set of seven direct-based keyframes $\kappa_{d}$ and
$2000$ active direct points. Increasing these parameters was found by \cite{engel_2016_ARXIV} to significantly increase the computational cost without much improvement in accuracy. 
Direct Keyframe insertion and marginalization occurs frequently
according to conditions described in \cite{engel_2016_ARXIV}. 
 In contrast, the feature-based map $M_{f}$
is made of an undetermined number of keyframes $\kappa_{f}$, each with an associated set of
features and their corresponding ORB descriptors $\Phi(x,Q(x))$.
\subsection{Odometry}
\subsubsection{Direct image alignment}
\begin{figure*}[!htb] 
\centering
\includegraphics[width=0.7\textwidth]{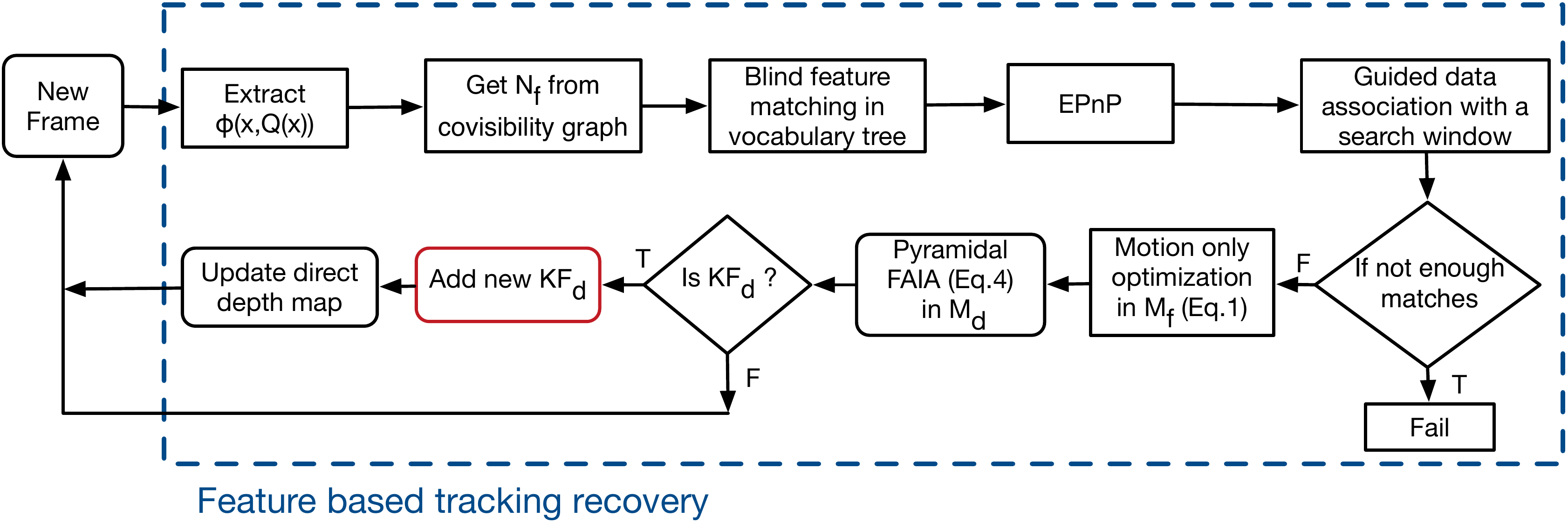}
\caption{FDMO Tracking Recovery flowchart. Only invoked when direct image
alignment fails, it takes over the front end operations of the system until the direct map is  re-initialized.  We start by extracting features from the new frame and matching them to 3D features observed in a set of keyframes $N_f$ connected to the last correctly added keyframe.  Efficient perspective n point (EPnP)  camera pose estimation is used to estimate an initial guess which is then refined by a guided data association between the local map and the frame. The refined pose is then used to seed a Forward additive image alignment step to achieve sub-pixel accuracy.}
\label{fig:feaTracking}
\end{figure*}

frame by frame operations are handled by the flowchart described in Fig.
\ref{fig:DirAlignment}. Similar to \cite{engel_2016_ARXIV}, newly acquired frames are tracked by minimizing \eqref{eq:FAIA} in
$M_{d}$, seeded from a constant velocity motion model (CVMM). However, erratic motion or large motion baselines can easily
violate the CVMM, erroneously initializing the highly-non convex optimization, and yielding
unrecoverable tracking failure. We detect tracking failure by monitoring the RMSE of the image alignment process before and after the optimization; if the ratio $\frac{RMSE_{after}}{RMSE_{before}}>1+\epsilon$ we consider that the optimization has diverged and we invoke the feature-based tracking recovery, summarized in the flowchart of Fig. \ref{fig:feaTracking}. The $\epsilon$ is used to restrict feature-based intervention when the original motion model used is accurate, a value of $\epsilon=0.1$ was found as a good trade-off between continuously invoking the feature-based tracking and not detecting failure in the optimization.
To avoid extra computational cost, feature extraction and matching is not performed on a frame by frame basis, it is only invoked during feature-based tracking recovery and feature-based KF insertion.

\subsubsection{Feature-based tracking recovery}
Our proposed feature-based tracking operates in $M_{f}$. When direct
tracking diverges, we consider the CVMM estimate to be invalid and seek to estimate a new motion model using the feature-based map. The new motion model is then used to re-initialize the direct image alignment. 
 Our proposed feature-based tracking recovery is a variant of the global re-localization method proposed in \cite{mur-artal_2015_TRO}; we first start by detecting FAST features with their associated ORB
descriptors $\Phi f_{f}=\Phi(x,Q(x))$ in the current image, which are then parsed into a vocabulary tree.  Since we consider the CVMM to be invalid, we fall back on the last piece of information the system was sure of before failure: the pose of the last successfully added keyframe. We define
a set of ten feature-based keyframes $N_{f}$ connected to the last added keyframe $KF_{d}$ through
a covisibility graph \cite{strasdat_2011_ICCV}, each with its associated $X\Phi KF_j$, where $KF_j\in N_{f}$, and
$X\Phi{KF_j}$ is the set of features from $KF_j$ that are associated with previously triangulated
map points.
Blind feature matching is then performed between $\Phi f_i$ and $\Phi KF_j $, by restricting feature
matching to take place between features that exist in the same node in a vocabulary tree \cite{lopez_2012_TRO}; this is done to reduce the computational cost of blindly matching all features.  

Once data association is established between $f_i$ and the map points observed in $N_{f}$, we set up an
EPnP (Efficient Perspective-n-Point Camera Pose Estimation) \cite{lepetit_2009_IJCV} to solve for an initial pose $T_{f_i}$ using 3D-2D correspondences in an non-iterative manner.
The new pose is then used to define a search window in $f_i$ surrounding the projected locations of all map points $X\in N_{f}$. 
Finally the pose $T_{f_i}$ is refined through the geometric optimization
defined by \eqref{eq:featTracking}.
To achieve sub-pixel accuracy, the recovered pose $T_{f_i}$ is then converted into a local increment over the pose of
the last active direct keyframe using $T_{f_i}\cdot T_{d,KF_{d}}$, and then further refined in a direct image alignment optimization \eqref{eq:FAIA}. 

Note that the EPnP step could have been skipped in favor of using the last correctly tracked keyframe's position as a starting point; however, it would require a larger search window, which in turn increases the computational burden of data association in the subsequent step; data association using a search window was also empirically found to fail when the baseline motion was relatively large.

\subsection{Mapping}
\begin{figure*}[!htb]
\centering
\includegraphics[width=0.7\textwidth]{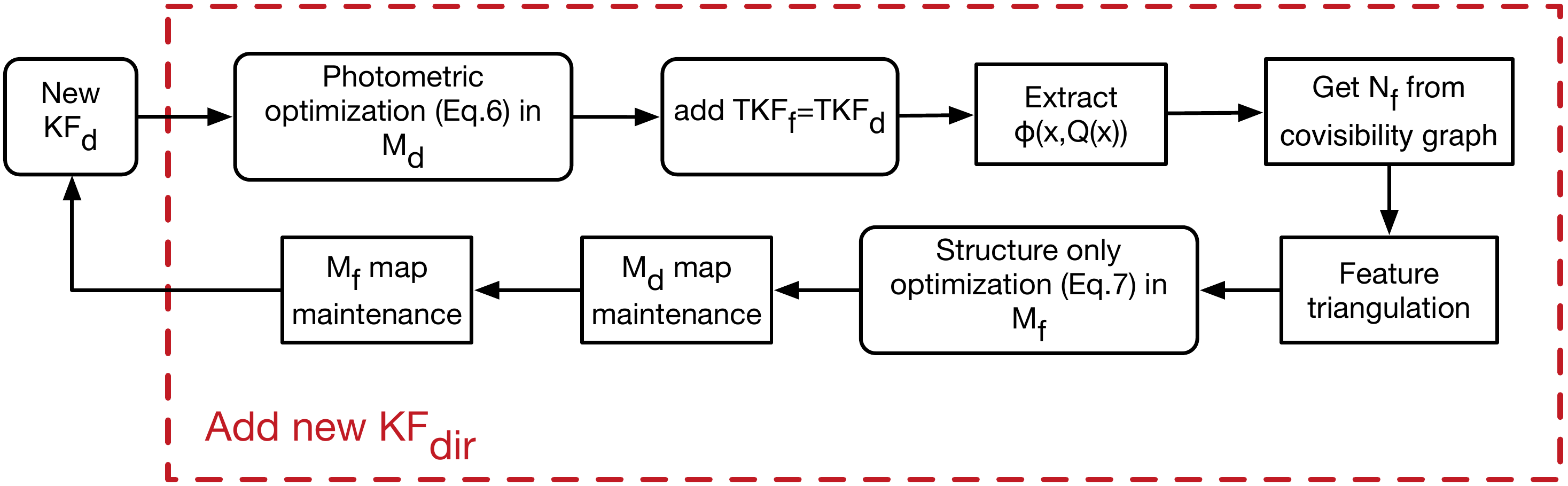}
\caption{Our proposed feature-based mapping flowchart; it operates after or parallel to the direct photometric
    optimization of \eqref{eq:PhBA}  and is responsible for expanding the feature-based map with
    new $KF_{f}$. It establish feature matches using restricted epipolar search lines; the 3D feature-based map is then optimized using a computationally efficient structure-only bundle adjustment, before map maintenance ensures the map remain outliers free .}
\label{fig:mapping}
\end{figure*}
The direct-feature-based map is expanded as described in Fig. \ref{fig:mapping}. When a new keyframe is added to $M_{d}$, we create a new feature-based keyframe $KF_{f}$ that inherits its
pose from $KF_{d}$ after its optimized through \eqref{eq:PhBA}. $\Phi KF_{f}(x,Q(x))$ is then
extracted and data association takes place between the new keyframe and a set of local keyframes
$\kappa'_{f}$ surrounding it via computationally efficient epipolar search lines.
The data association is used to keep track of all map points
$X_{f}$ visible in the new keyframe and to triangulate new map points.  

To ensure an accurate and reliable feature-based map, typical feature-based methods employ local bundle adjustment to optimize for both the keyframes poses and their associated map points; however, this is computationally very expensive and could severely reduce the frame rate of our proposed approach; instead, we make use of the fact that the new keyframe's pose
is locally optimal (optimized in the direct optimization of \eqref{eq:PhBA}), to replace the typical local bundle adjustment with a computationally less demanding structure-only optimization defined by:

\begin{equation}
	\label{eq:sba}
	\underset{X_j}{\operatorname{argmin}} \sum_{i\in\kappa'_{f}}\sum_{j} Obj(x_{i,j}-\pi(T_{KF_i},X_j)),
	\end{equation}
where $X_j$ spans all 3D map points observed in all keyframes $\in$ $\kappa'_{f}$.
 We limit the number of iterations in the optimization of \eqref{eq:sba} to ten, since no significant reduction in the feature-based re-projection error was recorded beyond ten iterations.

\subsection{Feature-based map maintenance}
To ensure a reliable feature-based map, the following practices are employed.
For proper operation, direct methods require frequent addition of keyframes, resulting in small baselines between the keyframes, which in turn can cause degeneracies if used to triangulate feature-based points. To avoid numerical instabilities, we prevent feature triangulation between keyframes with a $\frac{baseline}{depth}$ ratio less than the empirically tuned threshold of $0.02$ which is a trade-off between numerically unstable triangulated features and feature deprivation problems.
We exploit the frequent addition of keyframes as a feature quality check. In other words, a feature has to be correctly found in at least $4$ of the $7$ keyframes subsequent to the keyframe it was first observed in, otherwise it is considered spurious and is subsequently removed.
To ensure no feature deprivation occurs, a feature cannot be removed until at least 7 keyframes have been added since it was first observed. Finally, a keyframe with ninety percent of its points shared with other keyframes is removed from $M_{f}$ only once marginalized from $M_{d}$.

 The aforementioned practices ensure that sufficient reliable map points and features are available in the immediate surrounding of the current frame, and that only necessary map points and keyframes are kept once the camera moves on.

%% file: experiments.tex
\section{Experiments and Results}
\label{sec:exp}
To evaluate FDMO's tracking robustness, experiments were performed on several well-known
datasets \cite{euroc_2016} and \cite{mono_dataset}, and both qualitative and quantitative appraisal was conducted.  To further validate FDMO's effectiveness, the experiments were also repeated on state of the art open-source systems in both direct (DSO) and feature-based (ORB SLAM). For fairness  of comparison, we evaluate ORB SLAM as an odometry system (not as a SLAM system); therefore, similar to \cite{engel_2016_ARXIV} we disable its loop closure thread but we keep its global failure recovery, local, and global bundle adjustments intact. 
Note that we've also attempted to include results from SVO \cite{forster_2014_ICRA} but it continuously failed on most datasets, so we excluded it.
 
\subsection{Datasets}
\subsubsection{TUM MONO dataset} \cite{mono_dataset} contains 50 sequences of a camera moving along a path that begins and at ends at the same location. The dataset is photometrically calibrated: camera response function, exposure times and vignetting are all available; however, ground truth pose information is only available for two small segments at the beginning and end of each sequence; fortunately, such information is enough to compute translation, rotation, and scale drifts accumulated over the path, as described in \cite{mono_dataset}.

\subsubsection{EuRoC MAV dataset} \cite{euroc_2016} contains 11 sequences of stereo images recorded by a drone mounted camera. Ground truth pose for each frame is available from a Vicon motion capture system.

\subsection{Computational cost}
The experiments were conducted on an Intel Core i7-4710HQ 2.5GHZ CPU, 16 GB memory; no GPU acceleration was used. The time required by each of the processes was recorded and summarized in Table \ref{tab:Time}. Both DSO and ORB SLAM consist of two parallel components, a tracking process (at frame-rate\footnote{occur at every frame.}) and a mapping process (keyframe-rate\footnote{occur at new keyframes only.}). On the other hand, FDMO has three main processes: (1) a direct tracking process (frame-rate), (2) a direct mapping process (keyframe-rate), and (3) a feature-based mapping process (keyframe-rate). Both of FDMO's mapping processes can run either sequentially for a total computational cost of $200$ ms on a single thread, or in parallel on two threads. 
As Table \ref{tab:Time} shows, the mean tracking time for FDMO remains almost the same that of DSO: we don't extract features at frame-rate; feature based tracking in FDMO is only performed when the direct tracking diverges; the extra time is reflected in the slightly increased standard deviation of the computational time with respect to DSO. Nevertheless, it is considerably less than ORB SLAM's 23 ms.  As for FDMO's mapping processes, its direct part remains the same as DSO, whereas the feature-based part takes $153$ ms which is also significantly less than ORB SLAM's feature based mapping process that requires $236$ ms. 

\newcolumntype{M}[1]{>{\centering\arraybackslash}m{#1}}
\begin{table}[!htb]
	    	\renewcommand{\arraystretch}{1.1}

	\caption{Computational time (ms) for continuous processes in DSO, FDMO and ORB SLAM. (Empty means the system does not have the process.)}
	\label{tab:Time}
	\centering		

		\begin{tabular}{|c||M{1.3cm}||M{1.3cm}||M{1.45cm}|}
			\hline
			\textbf{Process} & \textbf{DSO} & \textbf{FDMO} & \textbf{ORB SLAM}\\
			\hline
			\parbox[c][0.7cm][c]{2.5cm}{\raggedright Tracking (frame-rate)}& 12.35$\pm$9.62 & 13.54$\pm$14.19& 23.04$\pm$4.11\\
			\hline
				\parbox[c][0.7cm][c]{2.5cm}{\raggedright Direct mapping (Keyframe-rate)}& 46.94$\pm$51.62 & 46.89$\pm$65.21& ---\\
			\hline
			\parbox[c][0.7cm][c]{2.7cm}{\raggedright Feature-based mapping (Keyframe-rate)}& --- & 153.8$\pm$58.08& 236.47$\pm$101.8\\
			\hline

		\end{tabular}
\vspace*{-0.5cm}
\end{table}

\subsection{Quantitative results}
\label{sec:qual_res}
We assess FDMO, ORB SLAM and DSO using the following experiments.
\subsubsection{Two loop experiment}
\begin{figure}[!htb]
 	\centering
 	\includegraphics[trim={0cm 0cm 0 0cm},clip,width=0.5\textwidth]{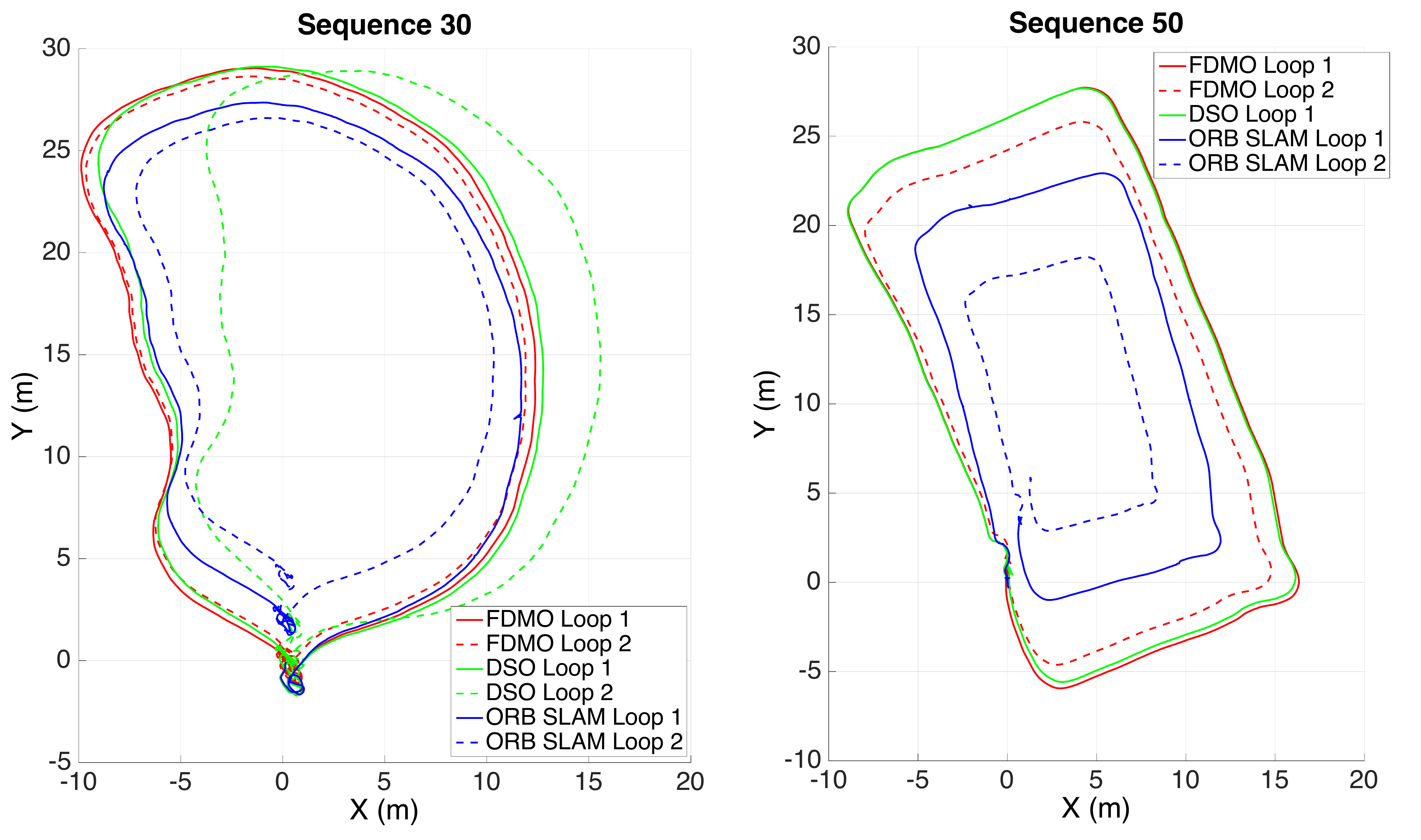}
 	\caption{Sample paths estimated by the various systems on Sequences 30 and 50 of the Tum\_Mono dataset. The paths are all aligned using ground truths available at the beginning and end of each loop. Each solid line corresponds to the first loop of a system while the dashed line correspond to the second loop. Ideally, all systems would start and end at the same location, while reporting the same trajectories across the two loops. Note that in Sequence 50, there is no second loop for DSO as it was not capable of dealing with the large baseline between the loops and failed. }
  	\label{fig:Traj}
 \end{figure}

in this experiment, we investigate the quality of the estimated trajectory by comparing ORB SLAM, DSO, and FDMO. We allow all three systems to run on various sequences of the Tum\_Mono dataset \cite{mono_dataset} across various conditions, both indoors and outdoors. Each system is allowed to run through every sequence for two continuous loops where each sequence begins and ends at the same location.
We record the positional, rotational, and scale drifts at the end of each loop, as described in \cite{mono_dataset}. The drifts recorded at the end of the first loop are indicative of the system's performance across that loop, whereas the drifts recorded at the end of the second loop consist of three components: (1) the drift accumulated from the first loop, (2)  an added drift accumulated over the second run, and (3) an error caused by a large baseline motion induced at the transition between the loops.
The reported results are shown in Table. \ref{tab:LoopExp} and some of the recovered trajectories are shown in Fig. \ref{fig:Traj}.
 \begin{table*}[!htb]
 	 \setlength{\abovecaptionskip}{-10pt}
 		\centering
 				\caption{Measured drifts after finishing one and two loops over various sequences from the TumMono dataset. The alignment drift (meters), rotation drift (degrees) and scale($\frac{m}{m}$) drifts are computed similar to \cite{mono_dataset}.}
 				\label{tab:LoopExp}
 		\includegraphics[trim={0cm 12cm 0 1cm},clip,width=\textwidth]{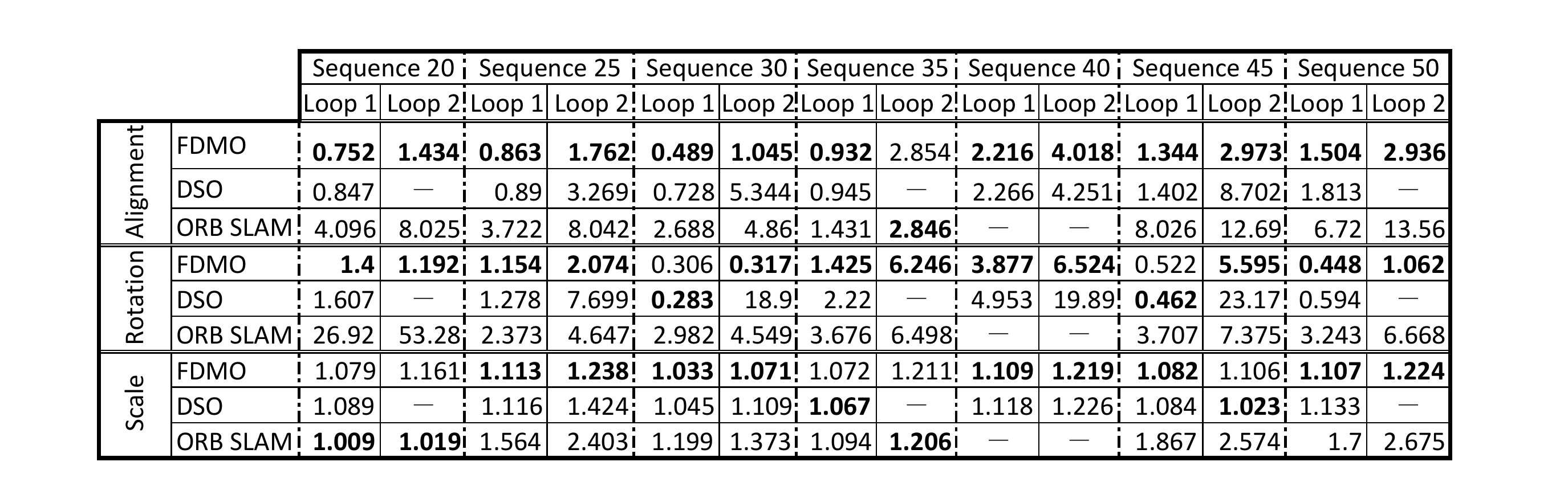}
 		\vspace*{-0.7cm}
\end{table*}

\subsubsection{Frame drop experiment}
\begin{figure}[!htb]
 	 \setlength{\belowcaptionskip}{-10pt}
 		\centering
 		\includegraphics[width=0.45\textwidth]{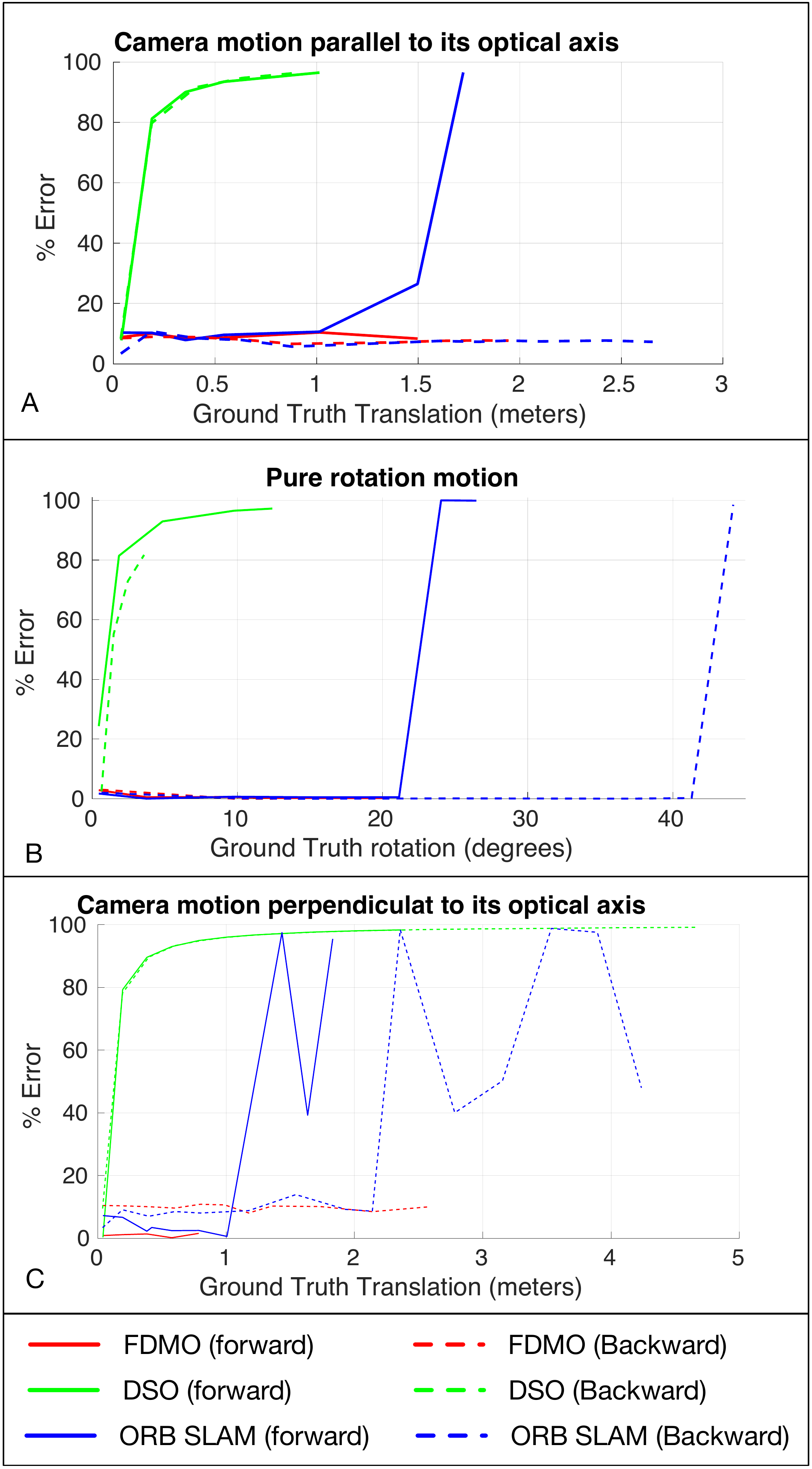}
 		\caption{$\% Error$ v.s. ground truth motion measured by dropping frames and estimating the relative transformation (rotation and translation) before and after the frames were dropped. (A) was conducted in the sequence MH01, (B) in the sequence MH02, and (C) in the sequence MH03 of the EuroC dataset \cite{euroc_2016}.   }
 		\label{fig:jumpexp}
 \end{figure}

While the first experiment reports on the  system's performance across large scale scenes in various conditions, this experiment investigates the effects erratic and large baseline motions have on the camera's tracking accuracy.
 Erratic motion can be defined as a sudden acceleration in the opposite direction of motion, and is quite common in hand-held devices or quad-copters. Another example of erratic motion occurs when the camera's video feed is being transmitted over a network to a ground station where computation is taking place; communication issues may cause frame drops which are seen by the odometry system as large baseline motions; therefore it is imperative for an odometry system to cope with such motions.
To quantize the influence of erratic motions on an odometry system, we set up an experiment to emulate their effects, by dropping a number of frames and measuring the recovered poses before and after dropping them. The experiment is repeated at the same location and the number of frames dropped is increased until each system fails. Various factors can affect the obtained results, such as the distance to the observed scene, skipping frames towards a previously observed or unobserved scene, and/or the type of camera motion (\textit{i.e.}, sideways, forward moving, or rotational motion), to name a few. Therefore we repeat the above experiment for each system in various locations covering the above scenarios.
We chose to perform the experiments on the EuroC dataset \cite{euroc_2016} whose frame to frame ground truth is known; thus allowing us to compute the relative Euclidean distance $Translation=||F_i-F_j ||$, and the orientation difference between the recovered poses at $F_i$ and $F_j$ as the \textit{geodesic metric of the normalized quaternions on the unit sphere} defined by $Rotation=cos^{-1}(2|F_i\cdot F_j|^2-1)$.
We report on the percent error $\% Error= 100 \times \frac{|Measured-GroundTruth|}{GroundTruth}$ for the recovered Euclidean distance and relative orientation before and after the skipped frames.
The obtained results for FDMO, DSO and ORB SLAM are shown in Fig. \ref{fig:jumpexp}. 

\subsection{Qualitative assessment}
\begin{figure}[!htb]
\centering
	 \setlength{\belowcaptionskip}{-15pt}
\includegraphics[trim={0cm 0cm 0 0cm},clip, width=0.46\textwidth]{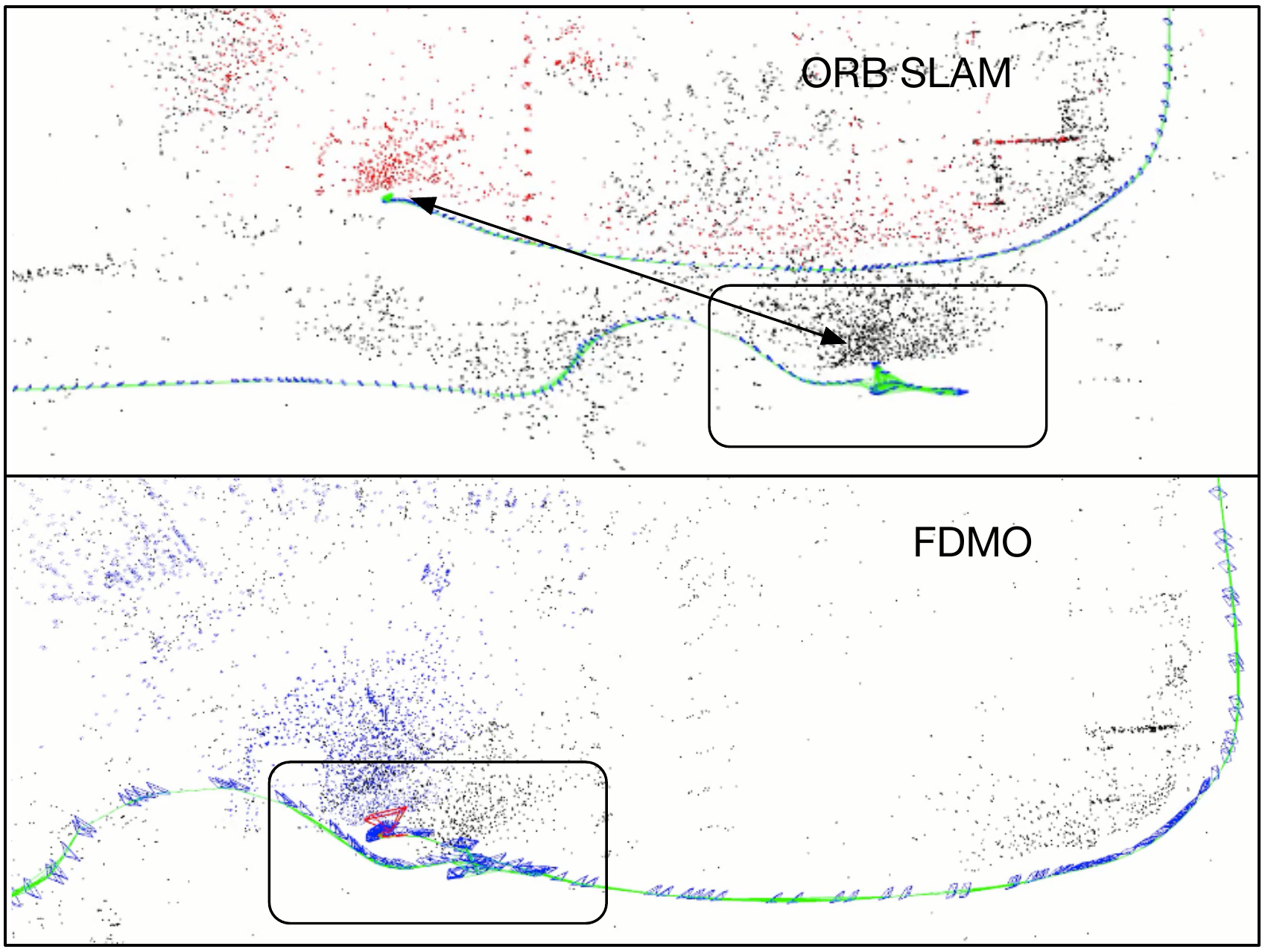}
\caption{Trajectory and feature-based maps estimated by ORB SLAM and FDMO after traversing Sequence 50 of the Tum\_mono dataset. }
\label{fig:quali}
\end{figure}

Fig. \ref{fig:quali} compares the feature-based map generated by FDMO to that of ORB SLAM (without loop closure). Notice the difference in the accumulated drift between both maps; FDMO's feature-based map inherited the sub-pixel accuracy of direct methods and did not suffer from severe drift. 
\subsection{Discussion}
The results reported in the first experiment (Table. \ref{tab:LoopExp}) demonstrate FDMO's performance in large-scale indoor and outdoor environments. 
The importance of the problem FDMO attempts to address is highlighted by analyzing the drifts incurred at the end of the first loop; while no artificial erratic motions nor large baselines were introduced over the first loop, FDMO was able to outperform both DSO and ORB SLAM in terms of positional, rotational, and scale drifts on most sequences. The improved performance is due to FDMO's ability to detect and account for inaccuracies in the direct framework using its feature-based map, while benefiting from the sub-pixel accuracy of the direct framework. 
Furthermore, FDMO was capable of expanding both its direct and feature-based maps in feature-deprived environments (\textit{e.g.} Sequence 40) whereas ORB SLAM failed to do so. FDMO's  robustness is further proven by analyzing the results obtained over the second loop. The drifts accumulated towards the end of the second loop are made of three components; mainly, the drift occurred over the first loop, the drift occurred over the second, and an error caused by a large baseline separating the frames at the transition between the loops. If the error caused by the large baseline is negligible, we would expect the drift at the second loop to be double that of the first. While the measured drifts for both ORB SLAM and FDMO does indeed exhibit such behavior, the drifts reported by ORB SLAM are significantly larger than the ones reported by FDMO as Fig. \ref{fig:Traj} also highlights. On the other hand,  DSO tracking failed entirely on various occasions, and when it did not fail, it reported a significantly large increase in drifts over the second loop. As DSO went through the transition frames between the loops its motion model estimate was violated, erroneously initializing its highly non-convex tracking optimization. The optimization got subsequently stuck in a local minimum, which led to a wrong pose estimate. The wrong pose estimate was in turn used to propagate the map, thereby causing large drifts. On the other hand, FDMO was successfully capable of handling such a scenario. 

The results reported in the second experiment (Fig. \ref{fig:jumpexp}) quantify the robustness limits of each system to erratic motions. Various factors may affect the obtained results, therefore, we attempted the experiments under various types of motion and by skipping frames towards a previously observed (herein referred to as backward) and previously unobserved part of the scene (referred to as forward). The observed depth of the scene is also an important factor: far-away scenes remain for a longer time in the field of view, thus improving the systems' performance. However, we cannot model all different possibilities of depth variations; therefore, for the sake of comparison, all systems were subjected to the same frame drops at the same locations in each experiment where the observed scene's depth varied from three to eight meters. The reported results highlight DSO's brittleness to any violation of its motion model; where translations as little as thirty centimeters and rotations as small as three degrees introduced errors of over $50\%$ in its pose estimates. On the other hand, FDMO was capable of accurately handling baselines as large as $1.5$ meters and $20$ degrees towards previously unobserved scene, after which failure occurred due to feature-deprivation, and two meters towards previously observed parts of the scene. ORB SLAM's performance was very similar to FDMO in forward jumps, however it significantly outperformed it by twice as much in the backward jumps; ORB SLAM uses a global map for failure recovery whereas FDMO, being an odometry system, can only make use of its immediate surroundings. Nevertheless FDMO's current limitations in this regard are purely due to our current implementation as there are no theoretical limitations of developing FDMO into a full SLAM system. However, using a global relocalization method has its downside; the jitter in ORB SLAM's behavior (shown in Fig. \ref{fig:jumpexp} (C)) is due to its relocalization process erroneously localizing the frame at spurious locations.
Another key aspect of FDMO visible in the this experiment, is its ability to detect failure and not incorporate it into its map. In contrast, towards their failure limits, both DSO and ORB SLAM incorporate spurious measurements for few frames before failing completely.



%% file: conclusions.tex
\section{Conclusion}
\label{sec:conc}
This paper successfully demonstrated the advantages of integrating direct and feature-based
methods in VO. By relying on a feature based map when direct tracking fails, the
issue of large baselines that is characteristic of direct methods is mitigated, while maintaining the high accuracy of
direct methods in both feature based and direct maps, and at a relatively low computational cost. Both qualitative and quantitative experimental results proved the effectiveness of the collaboration between direct and feature-based methods in the localization part.

While these results are exciting, they do not make use of a global feature based map; as such we are currently developing a more elaborate integration between both frameworks, to improve the mapping accuracy.  Furthermore, we anticipate that the benefits to the
mapping thread will also lead to added robustness and accuracy to the motion estimation within a full SLAM framework.